\documentclass[letterpaper, 10 pt, conference]{ieeeconf}  

\IEEEoverridecommandlockouts                              

\usepackage{cite}
\usepackage{tabularx}
\usepackage{multirow}
\usepackage{amsmath,amssymb,amsfonts}
\usepackage{graphicx}
\usepackage{textcomp}

\usepackage[table,xcdraw]{xcolor}
\usepackage[linesnumbered,ruled,lined]{algorithm2e}
\usepackage{booktabs}
\usepackage{xcolor}
\usepackage{authblk}
\let\labelindent\relax
\usepackage{enumitem}
\usepackage{hyperref}
\usepackage{breqn}
\usepackage{algpseudocode}
\usepackage{caption}
\usepackage{subcaption}
\usepackage{tcolorbox}
\usepackage{array}
\usepackage{makecell}
\usepackage{cuted}
\usepackage{adjustbox}
\usepackage{mdframed}

\newenvironment{tightalgorithm}
  {%
   \setlength{\textfloatsep}{5pt}%
   \setlength{\intextsep}{5pt}%
   \begin{algorithm}
  }
  {%
   \end{algorithm}
  }

\newcounter{parnum}[subsection] 
\renewcommand{\theparnum}{\alph{parnum}} 
\newcommand{\parletter}[1]{%
  \refstepcounter{parnum}%
  \noindent\textbf{\theparnum) #1}\par
}

\title{\LARGE \bf
Sampling-Based Optimization with Parallelized Physics Simulator for Bimanual Manipulation.
}
\author{Iryna Hurova$^{1}$, Alinjar Dan$^{1}$, Karl Kruusamäe$^{1}$, Arun Kumar Singh$^{1}$
\thanks{$^{1}$ University of Tartu, Narva mnt 18, 51009, Tartu, Estonia\\
\vspace{1mm}
Iryna Hurova: Post-graduate student, \url{iryna.gurova@gmail.com}\\
Alinjar Dan: Post-doc, \url{alinjardannitdgp2014@gmail.com}\\
Karl Kruusamäe: Associate Professor, \url{karl.kruusamae@ut.ee}\\
Arun Kumar Singh: Associate Professor, \url{aks1812@gmail.com}}
}

\makeatletter
\let\@oldmaketitle\@maketitle
\renewcommand{\@maketitle}{\@oldmaketitle
\centering
\includegraphics[width=0.988\linewidth]{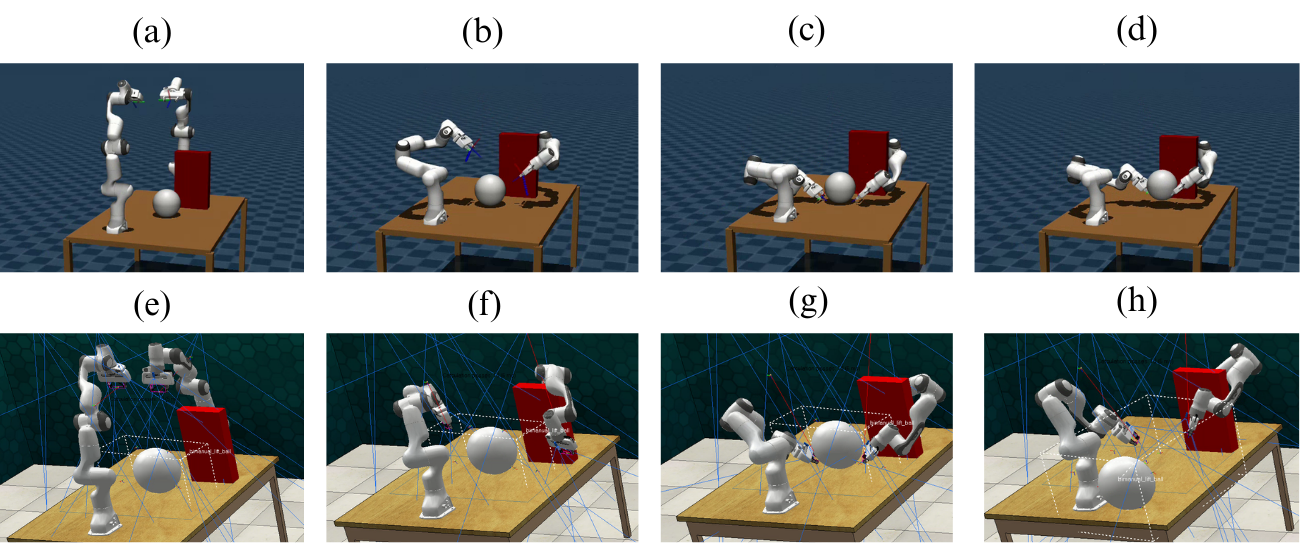}
\captionof{figure}{\footnotesize{Comparison of our approach (top) with \cite{gkanatsios20253d} (bottom). We consider the task of lifting the ball made complicated by the presence of an obstacle. In free space, both our approach and \cite{gkanatsios20253d} work flawlessly. However, \cite{gkanatsios20253d} is not able to adapt the lifting strategy in the presence of obstacles.}}
\label{fig:teaser}
\vspace{-0.3cm}
}
\makeatother

\usepackage{graphicx}

\begin{document}

\maketitle
\thispagestyle{empty}
\pagestyle{empty}

\begin{abstract}
In recent years, dual-arm manipulation has become an area of strong interest in robotics, with end-to-end learning emerging as the predominant strategy for solving bimanual tasks. A critical limitation of such learning-based approaches, however, is their difficulty in generalizing to novel scenarios, especially within cluttered environments. This paper presents an alternative paradigm: a sampling-based optimization framework that utilizes a GPU-accelerated physics simulator as its world model. We demonstrate that this approach can solve complex bimanual manipulation tasks in the presence of static obstacles. Our contribution is a customized Model Predictive Path Integral Control (MPPI) algorithm, \textbf{guided by carefully designed task-specific cost functions,} that uses GPU-accelerated MuJoCo for efficiently evaluating robot-object interaction. We apply this method to solve significantly more challenging versions of tasks from the PerAct$^{2}$ benchmark, such as requiring the point-to-point transfer of a ball through an obstacle course. Furthermore, we establish that our method achieves real-time performance on commodity GPUs and facilitates successful sim-to-real transfer by leveraging unique features within MuJoCo. The paper concludes with a statistical analysis of the sample complexity and robustness, quantifying the performance of our approach. The project website is available at: https://sites.google.com/view/bimanualakslabunitartu
\end{abstract}


\section{Introduction}
Dual-arm manipulation represents a significant frontier in robotics, holding the promise of emulating human-level dexterity for complex tasks such as intricate assembly, collaborative object handling, and dynamic interaction with unstructured environments. The ability to coordinate two robotic arms unlocks capabilities that are difficult or impossible for a single manipulator, including stabilizing large objects, performing simultaneous actions, and applying opposing forces. This potential has fueled a surge of research interest, with the goal of developing autonomous systems that can operate robustly and intelligently in the physical world.

In recent years, the dominant paradigm for tackling bimanual manipulation has been end-to-end learning~\cite{lu2024anybimanual, gkanatsios20253d, grotz2024peract2, black2024pi_0}. Approaches based on imitation learning and reinforcement learning have demonstrated remarkable success in learning complex, high-dimensional policies directly from data. There are two primary reasons for the popularity of learning based approaches. First, they can directly convert high-dimensional perception input to motor commands. Second, they bypass the need to explicitly model the interaction between the arms and between the object and the arms. For example, consider the task of lifting a ball with both arms. By directly learning from the demonstration data, imitation learning can avoid explicitly modeling the contact forces between the ball and the arms.  However, the success of these learning-based approaches is often predicated on extensive data collection and is tightly coupled to the distribution of the training data. A critical and widely acknowledged limitation is their difficulty in generalizing to novel scenarios, especially when the environment is cluttered with obstacles. When faced with even minor variations in object placement, environmental clutter, or task constraints not seen during training, these policies can become brittle and fail unpredictably (see Fig.~\ref{fig:teaser}). This brittleness severely hinders their deployment in real-world settings, which are inherently dynamic and cluttered.

This paper shows that real-time, model-based planning with a ground-truth world model can produce sophisticated bimanual behavior. As a result, it offers an alternative paradigm to circumvent the generalization challenges of pure learning-based methods, especially for contact tasks, where the manipulator-interaction physics pose a significant challenge. Furthermore, this model-based planning can be used to generate synthetic expert data using privileged simulator information \cite{chen2020learning}. We advocate for a sampling-based optimization framework that leverages a high-fidelity, GPU-accelerated physics simulator as a predictive world model. Instead of learning a monolithic policy, our approach uses the world model to rapidly simulate thousands of potential future action sequences in parallel, selecting the optimal one at each time step. This model-based planning approach allows the robot to react dynamically to the current state of the world, including the precise location of obstacles and objects, enabling robust performance in cluttered and novel situations.

At the core of our work is a customized Model Predictive Path Integral Control (MPPI) \cite{williams2017model} algorithm. Specifically, we build upon \cite{andrejev2025pi} and embed a Quadratic Programme (QP) within the MPPI sampling routine to induce a high degree of smoothness in the sampled trajectories and reduce the effect of covariance tuning. This controller is guided by carefully designed, task-specific cost functions that encode the desired behavior, and it harnesses the massively parallel processing power of modern GPUs to evaluate trajectory costs within a MuJoCo physics simulation. We demonstrate the efficacy of our method by solving significantly more challenging versions of tasks from the well-regarded PerAct$^{2}$ benchmark~\cite{grotz2024peract2}. For instance, we transform a simple ball-lifting task into a complex, long-horizon problem requiring the point-to-point transfer of a ball through a set of static obstacles. Our results show that the proposed model-based planning can be performed in real-time on commodity hardware, and we highlight a clear path to successful sim-to-real transfer by leveraging unique features within the MuJoCo simulator.

To summarize, our key contributions are:
\begin{enumerate}
    \item A \textbf{novel sampling-based optimizer} that can solve complex bimanual tasks.
    \item \textbf{A first-ever demonstration} of sampling-based optimization on more complicated variants of some of the PerAct$^{2}$ \cite{grotz2024peract2} benchmark tasks, which present difficulties for existing learning-based methods.
    \item \textbf{Evidence of real-time performance} on commodity GPUs and a successful sim-to-real transfer methodology, establishing the practical viability of our approach.
    \item \textbf{A rigorous statistical analysis} quantifying the sample complexity and robustness of our method, providing a clear understanding of its performance characteristics.
\end{enumerate}

\section{Related Works}
We divide the survey of existing works into two categories: model-based and model-free approaches leveraging end-to-end learning. \\
\subsection{Model-Based Planning} 
\noindent Gradient-based trajectory optimization has been used in the past for bimanual tasks such as synchronous pick-and-place \cite{martinez2025trajectory} and object transportation \cite{karim2025vil}, \cite{hu2022adaptive}. However, these methods are specifically designed for a given task, and it is difficult to generalize them to a more complex setting. For example, the extension of \cite{karim2025vil} to handle obstacles would require a major overhaul. Similarly, gradient-based optimization is not suitable for tasks involving the making or breaking of contacts. In contrast, approaches based on sampling-based optimization \cite{howell2022predictive}, \cite{pezzato2025sampling} can handle arbitrary tasks as long as it is possible to tractably rollout the environment/world-model for a given sequence of controls. Authors in \cite{tong2025adaptive} provide a way of improving sampling-based approaches. They learn an energy model that allows generating samples that respect different kinematic and collision constraints enforced by the bimanual setup. But \cite{tong2025adaptive} has not been applied to tasks such as ``ball-lifting" that involves making or breaking of contacts.
\subsubsection*{Our Improvement} Our approach is inspired by the preliminary bimanual results of \cite{howell2022predictive}. However, we consider a more powerful optimizer to handle a wider range of bimanual tasks, both in simulation and the real world. A unique feature of our approach is that the sampled velocity sequences are smooth(jerk-bounded). Our approach is also several times faster than \cite{tong2025adaptive} and allows for real-time control.

\subsection{End-to-End Learning:} 
\noindent Approaches like \cite{gkanatsios20253d}, \cite{lu2024anybimanual},  \cite{grotz2024peract2} have shown remarkable success in bimanual tasks by learning directly from expert demonstrations. A key ingredient of imitation learning (IL) based approaches has been the policy representation in terms of diffusion \cite{janner2022planning} or flow-matching models \cite{lipman2022flow}. However, there are a few core challenges that hinder the scalability of IL approaches. First, obtaining expert demonstration is challenging, especially in cluttered environments. Second, it is well known that neural network predictions typically struggle to satisfy constraints \cite{dontidc3}. Finally, the best performing ILs, based on diffusion or flow-matching, have high inferencing times \cite{gkanatsios20253d}.
The expert data problem, in principle, can be solved by reinforcement learning RL. However, the current approaches are either designed for specific tasks \cite{cui2024task} or consider simplified manipulator models \cite{chen2023bi}.
\subsubsection*{Our Improvement} Our approach can handle constraints such as collision avoidance and can run between $10-16 Hz$ even on a laptop GPU. Our approach can adapt to arbitrary tasks as long as they can be encoded by a well-defined cost function. Finally, our optimizer can also be crucial in adapting RL approaches like \cite{hansen2022temporal} to bimanual manipulation.

\section{Methods}
Consider a bimanual setting where $\boldsymbol{\theta}_{1, k}, \boldsymbol{\theta}_{2, k} $ represent the joint position at time-step $k$ for the individual arms. Let $\boldsymbol{\theta}_k=(\boldsymbol{\theta}_{1, k}, \boldsymbol{\theta}_{2, k})$ represents the combined joint position. Let $\mathbf{x}_k$ represent the environment state at step $k$. For example, $\mathbf{x}_k$ can contain the Euclidean position of any specific point on the manipulator body, the position and velocity of any movable object in the environment, etc. With these notations in place, we define bimanual planning over a horizon $H$ as the following optimization problem:
\begin{subequations}
 \begin{align}
    \arg\min_{\dot{\boldsymbol{\theta}}_{0:H} } \sum_k c(\boldsymbol{\theta}_k, \dot{\boldsymbol{\theta}}_k, \mathbf{x}_k ) \label{cost}\\
   \boldsymbol{\theta}_{k+1}, \mathbf{x}_{k+1} = \mathbf{f}(\boldsymbol{\theta}_k, \dot{\boldsymbol{\theta}}_{k}, \mathbf{x}_k) \label{state_evol} \\
   \boldsymbol{\theta}^{(r)}_{min} \leq \boldsymbol{\theta}_k^{(r)} \leq    \boldsymbol{\theta}^{(r)}_{max} \label{bounds}
\end{align}   
\end{subequations}

\noindent The cost function \eqref{cost} encodes the task requirements along with other constraints such as collision avoidance. We discuss the task-specific algebraic form of the costs in the later sections. The combined joint and task-space evolution is dictated by the function $\mathbf{f}$, which can be thought of as our world-model, defined implicitly as a MuJoCo physics engine. To ensure smoothness in the joint motions and aid in sim2real transfer, we enforce bounds on the $r^{th}$ derivative of the joint trajectory. In our implementation $r = \{0,1,2, 3\}$, i.e, we enforce constraints up to jerk level on the joint motions. It is worth pointing out that our optimization solves for both the arm motions together. Thus, it automatically discovers an appropriate coordination strategy in a given task. 

\subsection{Sampling Based Optimization}
\noindent Alg.\ref{algo_1} presents our sampling-based optimizer for solving \eqref{cost}-\eqref{bounds} built on top of MPPI. The process commences by initializing a Gaussian distribution, defined by a mean $\boldsymbol{\nu}_m$ and covariance $\boldsymbol{\Sigma}_m$, which represents the initial, broad search space for potential joint velocity inputs. Within a main loop that repeats for a set number of iterations, the optimizer systematically refines this search to find a high-quality solution. In each cycle, we first draw a large batch of $n$ candidate action sequences from its current probability distribution (line 6). Typically, these raw samples could be very noisy. Therefore, we project these samples onto the feasible set defined by the constraints \eqref{bounds} (line 7). This projection step can be formulated as QP \eqref{proj_cost}-\eqref{proj_bounds}, and it is solved for all the samples in parallel and accelerated over GPUs.  
\begin{align}
    \dot{\boldsymbol{\theta}} = \arg\min_{\dot{\boldsymbol{\theta}}} \frac{1}{2}\Vert \dot{\boldsymbol{\theta}}-\tilde{\dot{\boldsymbol{\theta}}}\Vert_2^2\label{proj_cost}\\
    \boldsymbol{\theta}^{(r)}_{min} \leq \boldsymbol{\theta}_k^{(r)} \leq    \boldsymbol{\theta}^{(r)}_{max} \label{proj_bounds}
\end{align}

With the valid and smooth joint trajectories obtained after projection, the algorithm then leverages a MuJoCo physics simulator to perform parallel ``rollouts", simulating the outcome of each velocity sequence to predict the resulting manipulator and environment states over time (line 8). Each of these simulated outcomes is subsequently evaluated using a cost function (line 9). In line 10, the algorithm identifies a small subset of the top-performing sequences, known as the \textbf{Elite Set}, based on their low costs. This elite set is then used to update the parameters of the Gaussian distribution through \eqref{mean_update}-\eqref{cov_update}, wherein $\eta$ represents the so-called learning rate and $\beta$ is the temperature parameter. 
\begin{subequations}
\begin{align}
    \boldsymbol{\nu}_{m+1} = (1-\eta)\boldsymbol{\nu}_m+\eta\frac{\sum_{j=1}^{j=n_{e}} s_j \dot{\boldsymbol{\theta}}_j   }{\sum_{j=1}^{j=n_{e}} s_j}, \label{mean_update}\\
    \boldsymbol{\Sigma}_{m+1} = (1-\eta)\boldsymbol{\Sigma}_m+\eta\Delta \boldsymbol{\Sigma}\label{cov_update}\\
   \Delta \boldsymbol{\Sigma} = \frac{ \sum_{j=1}^{j=n_{e}} s_j(\dot{\boldsymbol{\theta}}_j-\boldsymbol{\nu}_{m+1})(\dot{\boldsymbol{\theta}}_j-\boldsymbol{\nu}_{m+1})^T}   {\sum_{s=1}^{s=n_e} s_j} \\
    s_j = \exp \left(\frac{-1}{\beta}c_j\right)    \label{s_formula}
\end{align}
\end{subequations}
This cycle of sampling, projecting, simulating, evaluating, and updating continues, progressively converging on an optimal region of the solution space. Upon completion of all iterations, the algorithm outputs the single best joint velocity sequence corresponding to the lowest cost in the final elite set along with the final $\boldsymbol{\nu}_M, \boldsymbol{\Sigma}_M$. We average the first few steps of the optimal joint velocity and execute it on the manipulators before invoking the next re-planning. Moreover, we warm-start the next planning cycle with  $\boldsymbol{\nu}_M, \boldsymbol{\Sigma}_M$.

\noindent 
\begin{tightalgorithm}[!t]
\caption{Sampling-Based Optimizer to Solve \eqref{cost}-\eqref{bounds}}
\label{algo_1}
\small
\SetAlgoLined
$M$ = Maximum number of iterations\\
Input: MuJoCo world model $\mathbf{f}$.\\
Initiate mean $\boldsymbol{\nu}_m, \boldsymbol{\Sigma}_m$, at iteration $m=0$ for sampling control inputs $\dot{\boldsymbol{\theta}}$  \\
\For{$m=1, m \leq M, m++$}
{
     \vspace{0.1cm}
    Initialize $CostList$ = []\\
     \vspace{0.1cm}
    
    Draw batch of ${n}$ bimanual joint velocity sequences $(\tilde{\dot{\boldsymbol{\theta}}}_1, \tilde{\dot{\boldsymbol{\theta}}}_2, \tilde{\dot{\boldsymbol{\theta}}}_j, ...., \tilde{\dot{\boldsymbol{\theta}}}_n)$ from $\mathcal{N}(\boldsymbol{\nu}_m, \boldsymbol{\Sigma}_m)$ \\
    
    QP based smoothing $\forall j$ ${\dot{\boldsymbol{\theta}}}_j = \text{QP}(\tilde{\dot{\boldsymbol{\theta}}}_j )$ \tcp*[f]{\textcolor{blue}{Using a QP based projection to ensure smoothness in the sampled trajectories}}\\
    \vspace{0.1cm}

    Compute $n$ joint and state trajectory rollouts $\boldsymbol{\theta}_{k+1, j}, \mathbf{x}_{k+1, j} =\mathbf{f} (\boldsymbol{\theta}_{k, j}, \dot{\boldsymbol{\theta}}_{k, j})$  \tcp*[f]{\textcolor{blue}{Parallely simulate the joint velocties to obtain the joint position of the arms and the combined task-space state of the arm and the environment. }}\\  
    \vspace{0.1cm}
    
    
    
    
    
    
    \vspace{0.1cm}
    
    Define $c_j = \sum_k c(\boldsymbol{\theta}_{k, j}, \dot{\boldsymbol{\theta}}_{k, j}, \mathbf{x}_{k, j} )$
    

    
    append each computed $c_j$ to $CostList$ \\
    \vspace{0.1cm}
        
    $EliteSet  \gets$ Select top $n_{e}$ samples of ($\dot{\boldsymbol{\theta}}$) with lowest cost from $CostList$.\\
     \vspace{0.1cm}
    ($\boldsymbol{\nu}_{m+1}, \boldsymbol{\Sigma}_{m+1} ) \gets$ Update distribution based on $EliteSet$ 
}
\Return{ Joint velocity $\dot{\boldsymbol{\theta}}$ corresponding to the lowest cost in the $EliteSet$ and $\boldsymbol{\nu}_{M}, \boldsymbol{\Sigma}_{M}$}
\normalsize
\end{tightalgorithm}
\vspace{-0.5cm}
\subsection{Role of QP}
\noindent The Quadratic Program (QP) of \eqref{proj_cost}-\eqref{proj_bounds} serves a dual purpose that goes beyond simply smoothing velocity samples from a Gaussian distribution. Critically, it eases the task of initializing the covariance matrix. This allows us to set a high noise level in the covariance matrix to encourage exploration, relying on the QP to find the optimal jerk-bounded joint trajectory that best fits the sampled velocity sequence. Consequently, the QP facilitates robust exploration while preventing oscillatory manipulator motions. This approach is well-supported, as previous studies have shown that smoothness in the sampled sequences and their resulting optimal controls enhances the overall performance of the sampling optimizers such as MPPI \cite{andrejev2025pi}, \cite{vlahov2024low}.

\begin{figure}
    \setcounter{figure}{1}
    \centering
    \includegraphics[width=0.4\textwidth]{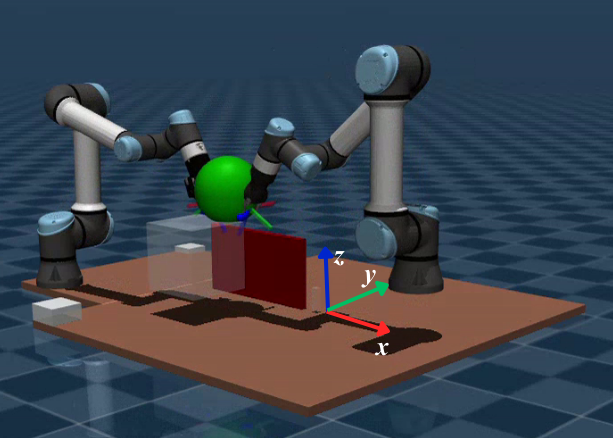}
    \caption{\footnotesize{Manipulator Arms in global frame.}}
    \label{fig:robot_global_frame}
    \vspace{-0.6cm}
\end{figure}
\section{Applications}
This section presents one of the main results of our paper. We consider more difficult versions of three tasks from the PerAct$^2$ benchmark and derive cost functions that can guide the sampling-based optimizer of Alg.\ref{algo_1} towards task completion.

\subsection{Moving tray}\label{subsec:Issue 19: moving tray}

\noindent This is a more complicated version of the tray-lifting task of the PerAct$^2$ benchmark \cite{grotz2024peract2}. Here, the tray needs to be lifted and transported to a given goal pose. We divide the task into two phases: the \textbf{pick} and \textbf{move} phases. The former pertains to guiding the manipulators to the grasp location, while the latter is designed for the transportation of the tray. Let the end-effector pose of $i$-th manipulator be denoted as $\mathbf{e}_{i, k} = (\mathbf{p}_{i, k}, \mathbf{q}_{i, k}) $, where $\mathbf{p}_{i, k}$ and $\mathbf{q}_{i, k}$ are respectively the position and orientation (as unit quaternions) of the manipulator and $i = \{1, 2\}$. Note that $\mathbf{e}_{i, k}$ is part of the augmented environment state $\mathbf{x}_{k}$ that is obtained by rolling out the velocity sequence on the MuJoCo simulator.



With these notations in place, we define the tray-moving task through the following cost functions.

\parletter{Collision Cost}
\noindent Let \( \mathbf{d}_k \in \mathbb{R}^{B} \) denote the concatenation of distances at time-step \( k \) between the two manipulators and \( B \) contact points. Collision avoidance is ensured if $d_k\geq 0, \forall k$. Define: 


\[
\mathbf{g}_k = -\mathbf{d}_{k+1} + (1 - \gamma) \mathbf{d}_k, \qquad \gamma \in (0, 1) 
\]

\noindent Then our proposed collision cost is:
\begin{align}
    \mathcal{C}_c = \sum_{k=0}^{H-2} \max( \mathbf{g}_{k}, 0 ) + \sum_{k=0}^{H-1} \mathbf{1}_{\{ \mathbf{d}_{k} < 0 \}}
    \label{collision_cost}
\end{align}

\noindent where \( \mathbf{1} \) is an indicator function, which is one if the condition in the parentheses holds and zero otherwise. The second term is the conventional collision cost that penalizes penetration between the manipulator body and the environment. The first term is the so-called barrier function inspired by \cite{zeng2021safety} that prevents $\mathbf{d}_k$ from reaching the boundary of the infeasible region.

\parletter{Joint Deviation Cost}
\noindent Let \( \boldsymbol{\theta}_{\text{home}} \) be a vector that denotes the home position of both the manipulators. We penalize deviation from this home position through the following cost.
\begin{align}
    \mathcal{C}_\theta = \sum_{k = 0}^{H-1}\left\| \boldsymbol{\theta}_k - \boldsymbol{\theta}_{\text{home}} \right\|_F
    \label{home_cost}
\end{align}

\noindent where \( \|\cdot\|_F \) denotes the Frobenius norm.\\
\parletter{End-effector Vertical Alignment Cost}
\noindent This cost penalizes deviations in the $z$ coordinates (see Fig.~\ref{fig:robot_global_frame}) of the two manipulators, ensuring they remain at the same height at every time step.
\begin{align}
    \mathcal{C}_z = \left\| \mathbf{p}_{z,1} - \mathbf{p}_{z,2} \right\|_2
    \label{vertical_alignment_cost}
\end{align}


\parletter{Relative Velocity Cost} 
\noindent Once the tray has been lifted, we want both the manipulators to move in a coordinated fashion to account for the fact that both the end-effectors are now connected to a rigid body. We model this requirement through the following cost:
\begin{align}
    \mathcal{C}_v = \sqrt{ \sum_{k=0}^{H-1} \left( \big(\mathbf{p}_{1,k} - \mathbf{p}_{2,k}\big) \cdot \big(\dot{\mathbf{p}}_{1,k} - \dot{\mathbf{p}}_{2,k}\big) \right)^2 }
    \label{rel_vel_cost}
\end{align}

\noindent This cost forces the relative velocity between the end-effectors to be perpendicular to their relative position, ensuring coordinated motion that avoids stretching or compressing, which is essential for stably carrying a rigid object.

\parletter{Pick Position Cost}
\noindent Let $\mathbf{p}_{\text{grasp, i}}$ denote the desired position at which the $i^{th}$ manipulator should grasp the tray for lifting. We define the following cost to guide the manipulators from their home position to the grasp position
\begin{align}
    \mathcal{C}_{p,\text{pick}} = \frac{1}{2} \sum_{k=0}^{H-1} \Big( \|\mathbf{p}_{1,k} - \mathbf{p}_{\text{grasp, 1}}\|_2 + \|\mathbf{p}_{2,k} - \mathbf{p}_{\text{grasp, 2}}\|_2 \Big)
    \label{tray_grasp_pos_cost}
\end{align}
\parletter{Pick Orientation Cost}

\noindent Let $\mathbf{q}_{\text{grasp, i}}$ denote the desired grasp orientation in the form of the unit quaternion. Then, the following cost is used to guide the manipulators to the right grasp orientation for lifting.
\begin{dmath}
    \mathcal{C}_{r,\text{pick}} = \frac{1}{2} \sum_{k=0}^{H-1} \left( 2 \cos^{-1} \left| \langle \mathbf{q}_{1,k}, \mathbf{q}_{\text{grasp,1}} \rangle \right| + 2 \cos^{-1} \left| \langle \mathbf{q}_{2,k}, \mathbf{q}_{\text{grasp,2}} \rangle \right| \right)
    \label{tray_grasp_ori_cost}
\end{dmath}

\parletter{End-effector Distance Cost}
\noindent Let $l_{\text{tray}}$ be the dimension of the tray along which it is held by the manipulators during lifting. The following cost ensures that both arms maintain this distance while moving.
\begin{align}
 \mathcal{C}_l = \sum_{k=0}^{H-1} \left( \| \mathbf{p}_{1,k} - \mathbf{p}_{2,k} \|_2 - l_{\text{tray}} \right)^2
 \label{end_effector_distance_cost}
\end{align}

\parletter{Tray Position Cost}  
\noindent Let $\mathbf{p}_{\text{tray}, k}$ be the position of the tray at time-step $k$. This is part of the augmented state vector $\mathbf{x}_k$ that is obtained through the physics simulator. We want the tray to move to a target position given by $\mathbf{p}_{\text{target}}$ and thus formulate the following cost.
\begin{align}
\mathcal{C}_{p,\text{tray}} = \sum_{k=0}^{H-1}\left\| \mathbf{p}_{\text{tray}, k} - \mathbf{p}_{\text{target}} \right\|_F
\label{tray_position_cost}
\end{align}



\parletter{Tray Orientation Cost}
\noindent In a similar manner, we formulate the following cost to place the tray in the orientation given by $\mathbf{q}_{\text{target}}$.
\begin{align}
    \mathcal{C}_{r,\text{tray}} = \sum_{k=0}^{H-1} 2 \cos^{-1} \left| \langle \mathbf{q}_{\text{tray},k}, \mathbf{q}_{\text{target}} \rangle \right|
    \label{tray_ori_cost}
\end{align}

\begin{mdframed}
      Note that the tray is not an actuated body. Thus, its position and orientation can only be indirectly controlled through the manipulators.
\end{mdframed}

\parletter{Move Orientation Cost}
\noindent During the \textbf{move} phase, we define some target end-effector orientations ($\mathbf{q}_{\text{move}, i}$) for each manipulator. This is done to indirectly control the orientation of the tray during motion.

\begin{align}
\mathcal{C}_{r,\text{move}} 
=& \tfrac{1}{2} \sum_{k=0}^{H-1} \Big(
    2 \cos^{-1}\!\big| \langle \mathbf{q}_{1,k}, \mathbf{q}_{\text{move}_{1,k}} \rangle \big| \notag \\[-4pt]
&+ \, 2 \cos^{-1}\!\big| \langle \mathbf{q}_{2,k}, \mathbf{q}_{\text{move}_{2,k}} \rangle \big|
\Big)
\label{tray_move_ori_cost}
\end{align}

\normalsize
\parletter{Total Cost}
The final cost for the tray-moving task is defined as :
\begin{align}
\mathcal{C} &= w_c \, \mathcal{C}_c 
+ w_\theta \, \mathcal{C}_\theta 
+ w_z \, \mathcal{C}_z 
+ w_v \, \mathcal{C}_v \notag \\
&\quad + w_{\text{pick}} \, \big( w_{p,\text{pick}} \, \mathcal{C}_{p,\text{pick}} + w_{r,\text{pick}} \, \mathcal{C}_{r,\text{pick}} \big) \notag \\
&\quad + w_{\text{move}} \, \big( w_l \, \mathcal{C}_l 
+ w_{p,\text{tray}} \, \mathcal{C}_{p,\text{tray}} \notag \\
&\quad \qquad \qquad + w_{r,\text{tray}} \, \mathcal{C}_{r,\text{tray}} 
+ w_{r,\text{move}} \, \mathcal{C}_{r,\text{move}} \big)
\end{align}
Please note $w_{\text{pick}}=1$, $w_{\text{move}}=0$ during \textbf{pick} phase and  $w_{\text{pick}}=0$, $w_{\text{move}}=1$ during the \textbf{move} phase. The other $w$'s are manually tuned.

\subsection{Lifting and Transporting ball}\label{subsec:Issue 21: lifting ball}

\noindent This task is an enhancement over the standard ball-lifting task of PerAct$^2$ benchmark \cite{grotz2024peract2}. In this task, both manipulator needs to cooperatively lift a ball by applying forces on the ball through their end-effector and subsequently transport it to a given location while avoiding obstacles. Similar to the earlier ``tray-moving", this task is divided into two phases: \textbf{pick} and \textbf{move}. The former is designed to position the manipulators to a location from where they can apply appropriate contact forces on the ball. The second phase is designed to ensure that the ball is not dropped during transportation.

\parletter{Collision Cost}
\noindent The collision cost is the same as \eqref{collision_cost} but calculated separately for \textbf{pick} and \textbf{move} phase as $\mathcal{C}_{c,\text{pick}}$ and $\mathcal{C}_{c,\text{move}}$ respectively. During the \textbf{move} phase, we remove the collision consideration between the ball and the manipulators. This is to ensure that the end-effectors are not repelled by the ball and instead press against its surface to apply contact forces.

\parletter{Joint Deviation Cost}
\noindent This is the same as \eqref{home_cost}.

\parletter{End-effector Planar Alignment Cost}
\noindent We want both the manipulators to apply contact forces on the ball at the same point in the $yz$ plane (Refer Fig.~\ref{fig:robot_global_frame}). Thus, we formulate the following alignment cost between the manipulators, where $\mathbf{p}_{iyz, k}$ represents the $yz$ component of $\mathbf{p}_i$
\begin{align}
    \mathcal{C}_{yz} = \sum_{k=0}^{H-1}\left\| \mathbf{p}_{1yz, k} - \mathbf{p}_{2yz, k} \right\|_F
    \label{planar_alignment_cost}
\end{align}


\parletter{Relative Velocity Cost}
\noindent This is the same as \eqref{rel_vel_cost}

\parletter{End-effector Orientation cost}
\noindent We define some target orientation at which the end-effector should come in contact with the ball. Subsequently, after lifting, the same end-effector orientation needs to be maintained during transporting the ball. Let \( \mathbf{q}^{\ast}_1, \mathbf{q}^{\ast}_2 \) be the target orientations in the form of unit quaternions, using which we formulate the following cost. 
\begin{align}
    \mathcal{C}_r = \tfrac{1}{2} \sum_{k=0}^{H-1} \Big( 2\cos^{-1}\big| \langle \mathbf{q}_{1,k}, \mathbf{q}^{\ast}_1 \rangle \big| + 2\cos^{-1}\big| \langle \mathbf{q}_{2,k}, \mathbf{q}^{\ast}_2 \rangle \big| \Big)
\end{align}

\vspace{-1em}

\parletter{End-effector Distance Cost}
\noindent This is the same as \eqref{end_effector_distance_cost} but with $l_{tray}$ replaced with the $l_{ball}$ which is the distance between the end-effector-ball contact points.
This cost ensures that a fixed offset of two arms along two points on the ball (not diametrically opposite) is maintained.


\parletter{End-effector to Object (ball) Alignment Cost}
\noindent Define the midpoint of the two end-effectors:
\[
\mathbf{c}_k = \tfrac{1}{2} \big(\mathbf{p}_{1,k} + \mathbf{p}_{2,k}\big).
\]
We penalize deviation from the ball target location (with small vertical offset) while \textbf{picking}:
\begin{align}
    \mathcal{C}_{\text{eef-obj}} = \sum_{k=0}^{H-1} \left\| \mathbf{c}_k - \big\{\mathbf{p}_{\text{ball}} - (0,0,\epsilon)\} \right\|_2.
    \label{ball_alignment_cost}
\end{align}
where $\epsilon$ is a small vertical offset.

\noindent where $\mathbf{p}_{\text{ball}}$ is the ball position before being picked.

\parletter{Object-to-Target Cost}
Let $\mathbf{p}_{\text{target}}$ be desired ball position. We formulate the following transportation cost for all
\begin{align}
    \mathcal{C}_{\text{obj-targ}} = \sum_{k=0}^{H-1} \| \mathbf{c}_k + (0,0,\epsilon) - \mathbf{p}_{\text{target}}\ \|_2.
\end{align}

\parletter{Final Total Cost}
\vspace{-1em}
\begin{align}
\mathcal{C} =& w_{\text{pick}} (w_c \, \mathcal{C}_{c,\text{pick}} + w_{\text{eef-obj}} \, \mathcal{C}_{\text{eef-obj}}) \notag \\
&+ w_{\text{move}} (w_c \, \mathcal{C}_{c,\text{move}} + w_{\text{obj-targ}} \, \mathcal{C}_{\text{obj-targ}}) + w_\theta \, \mathcal{C}_\theta \notag \\ 
&+ w_{yz} \, \mathcal{C}_{yz}
+ w_v \, \mathcal{C}_v
+ w_r \, \mathcal{C}_r 
+ w_l \, \mathcal{C}_l \notag \\
\end{align}
The weights $w_{\text{pick}}$ and $w_{\text{move}}$ are activated or deactivated in the same way as mentioned for the earlier task.

\subsection{Hand-Over Task}\label{subsec:Issue 22: pass cubes}
\noindent This task is again taken from the PerAct$^2$ benchmark \cite{grotz2024peract2} but made more complicated by the addition of static obstacles. We divide this task into three phases: \textbf{pick}, \textbf{pass}, and \textbf{place}.

\parletter{Collision Cost}
Same as \eqref{collision_cost}

\parletter{Joint Deviation Cost}
Same as \eqref{home_cost}



\parletter{End-effector Planar Alignment Cost}
\noindent While performing the handover in \textbf{pass} phase, we want both the manipulators to be aligned in the $yz$ plane. Thus, we use the cost defined in \eqref{planar_alignment_cost}.


\parletter{Phase-Specific Position Costs}
\noindent Let $\mathbf{p}_{\text{obj}}$, $\mathbf{p}_{\text{pass},i}$, and $\mathbf{p}_{\text{target}}$ be the desired positions for \textbf{pick}, \textbf{pass}, and \textbf{place} of the $i^{th}$ manipulator. For each arm $i$, we define
\begin{align}
\text{Pick:} \quad 
& \mathcal{C}_{p,i}^{\text{pick}} =
\sum_{k=0}^{H-1} \big\| \mathbf{p}_{i,k} - (\mathbf{p}_{\text{obj}} + [0,0,\epsilon]) \big\|_2, \\[4pt]
\text{Pass:} \quad 
& \mathcal{C}_{p,i}^{\text{pass}} =
\sum_{k=0}^{H-1} \big| \mathbf{p}_{(x)i,k} - \mathbf{p}_{(x)\text{pass},i} \big|, \\[4pt]
\text{Place:} \quad 
& \mathcal{C}_{p,i}^{\text{place}} =
\sum_{k=0}^{H-1} \big\| \mathbf{p}_{i,k} - \mathbf{p}_{\text{target}} \big\|_2.
\end{align}

The combined position cost is
\[
\mathcal{C}_p =
\sum_{i=1}^2 \Big(
w_{i}^{\text{pick}} \, \mathcal{C}_{p,i}^{\text{pick}} +
w_{i}^{\text{pass}} \, \mathcal{C}_{p,i}^{\text{pass}} +
w_{i}^{\text{place}} \, \mathcal{C}_{p,i}^{\text{place}}
\Big).
\]

\parletter{Phase-Specific Orientation Costs}
For each phase, i.e., (\textbf{pick}, \textbf{pass}, and \textbf{place}) define a target unit quaternion \(\mathbf{q}_{\text{target}}\).
The geodesic quaternion distance is
\[
\mathcal{C}_{r,i}^{\text{phase}} =
\sum_{k=0}^{H-1} 2 \arccos \Big(
\big| \langle \mathbf{q}_{i,k}, \, \mathbf{q}_{\text{target}} \rangle \big|
\Big).
\]

The combined orientation cost is
\[
\mathcal{C}_r =
\sum_{i=1}^2 \Big(
w_{i}^{\text{pick}} \, \mathcal{C}_{r,i}^{\text{pick}} +
w_{i}^{\text{pass}} \, \mathcal{C}_{r,i}^{\text{pass}} +
w_{i}^{\text{place}} \, \mathcal{C}_{r,i}^{\text{place}}
\Big).
\]

\parletter{Total Cost}
The overall cost is defined as
\begin{align}
\mathcal{C} =
w_c \, \mathcal{C}_c
+ w_{\theta} \, \mathcal{C}_\theta + w^{\text{pass}} w_{yz} \, \mathcal{C}_{yz} + w_p \, \mathcal{C}_p
+ w_r \, \mathcal{C}_r
\end{align}

\begin{figure}[!t]
    \centering
    \includegraphics[width=0.85\linewidth]{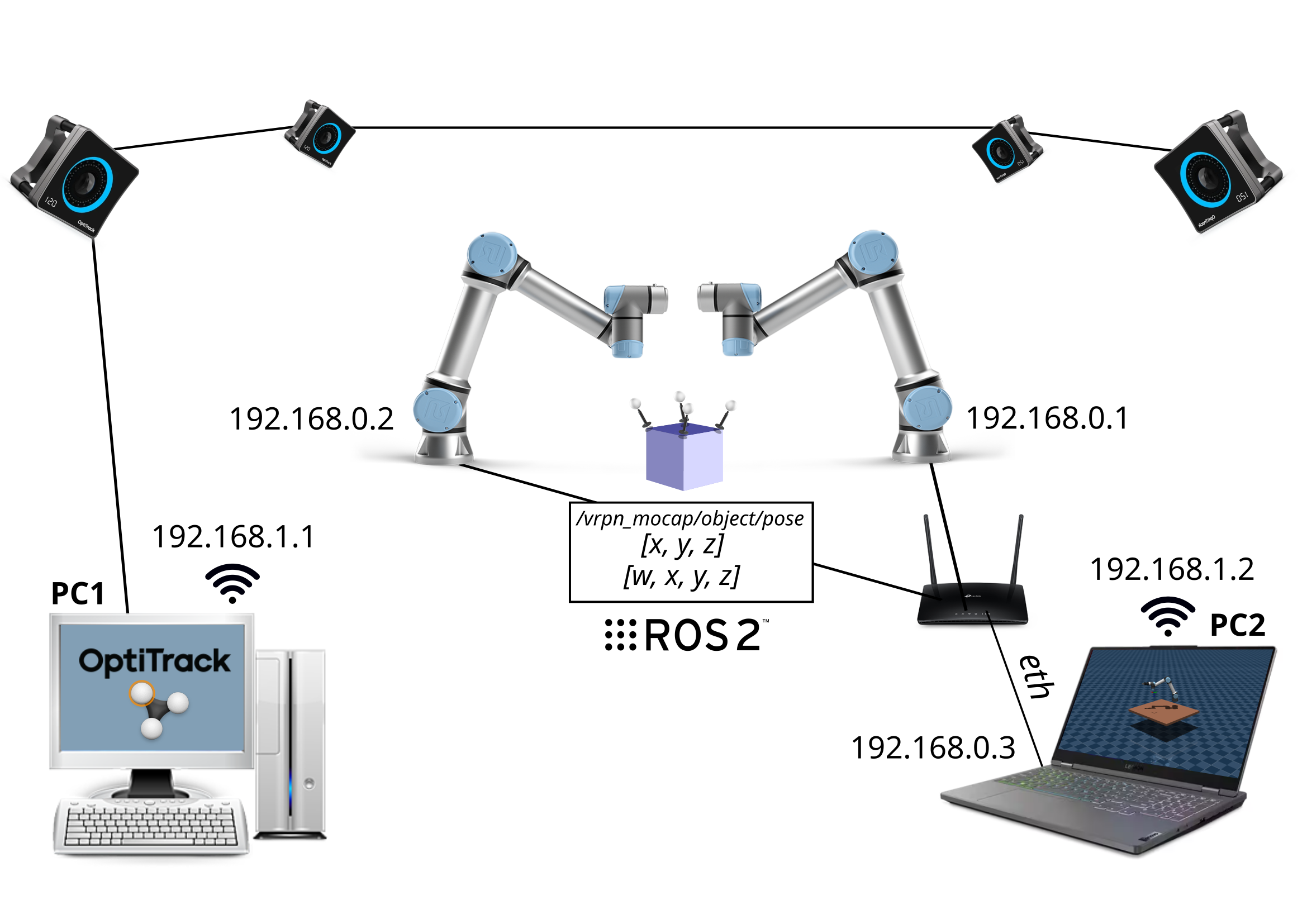}
    \caption{\footnotesize{Experimental Setup wherein real-world and simulation are tightly coupled with OptiTrack motion capture.}}
    \label{fig:real_life_setup}
    \vspace{-2em}
\end{figure}

\section{Simulation and Experimental Results}
\subsection{Implementation Details}
\noindent Alg.\ref{algo_1} was implemented in JAX with the GPU-accelerated MJX physics engine for receding horizon control. The simulation, featuring two UR5e manipulators with Robotiq grippers, uses a 0.1s time-step that was experimentally chosen to balance accuracy and planning time. We validate our approach with the following metrics.

\begin{figure*}
    \centering
    \includegraphics[width=\linewidth]{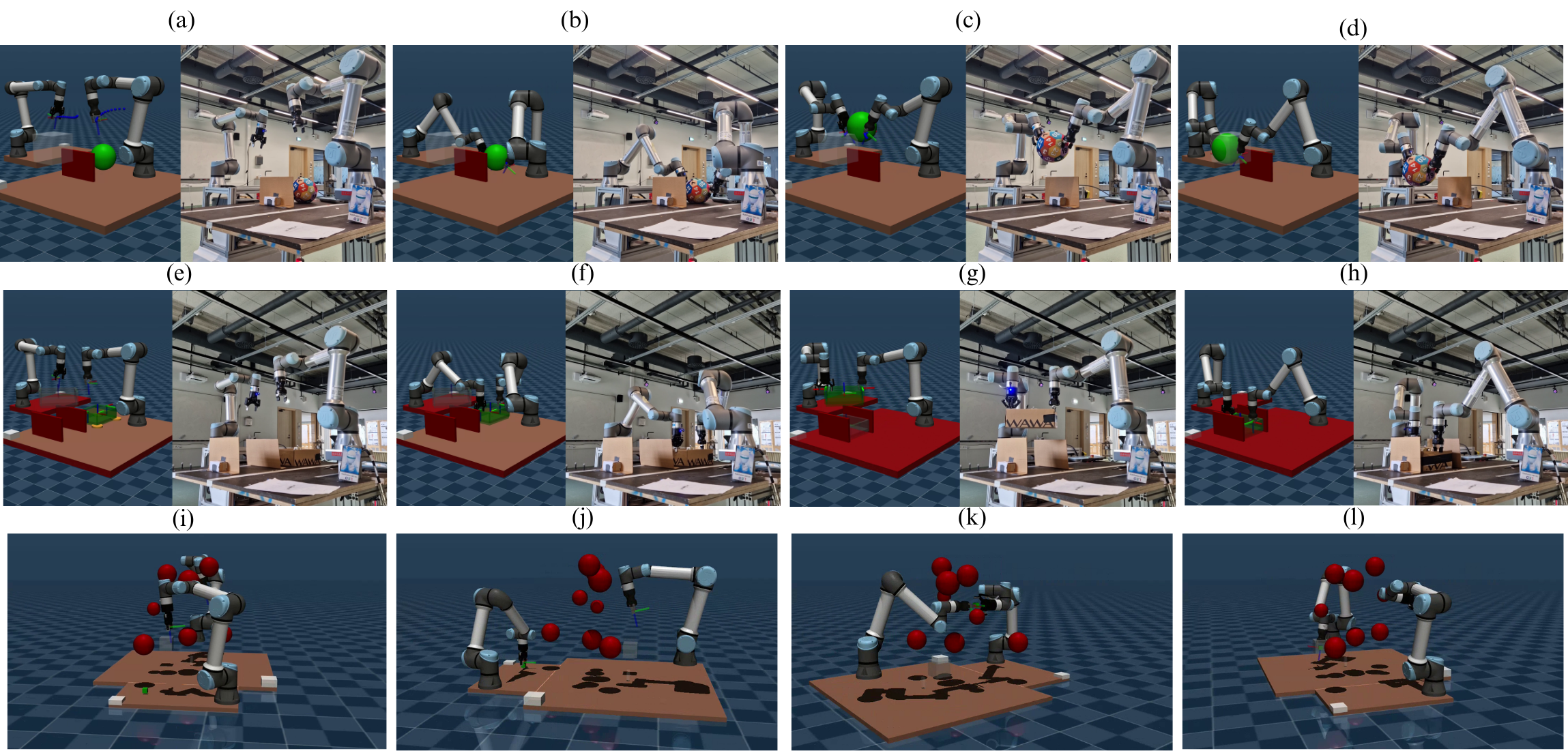}
    \caption{\footnotesize{Snapshot of all tasks. Top row: Manipulators lifting a ball while avoiding obstacles. Middle row: Manipulators picking a tray and placing it between two wall-like obstacles. Last row: Manipulators performing hand-over tasks amidst static obstacles. Note that the viewing angle for each snapshot is different. Refer to the accompanying videos for further details.}}
    \label{fig:all_task_snapshots}
\end{figure*}

\begin{figure*}
    \centering
    \includegraphics[width=\textwidth]{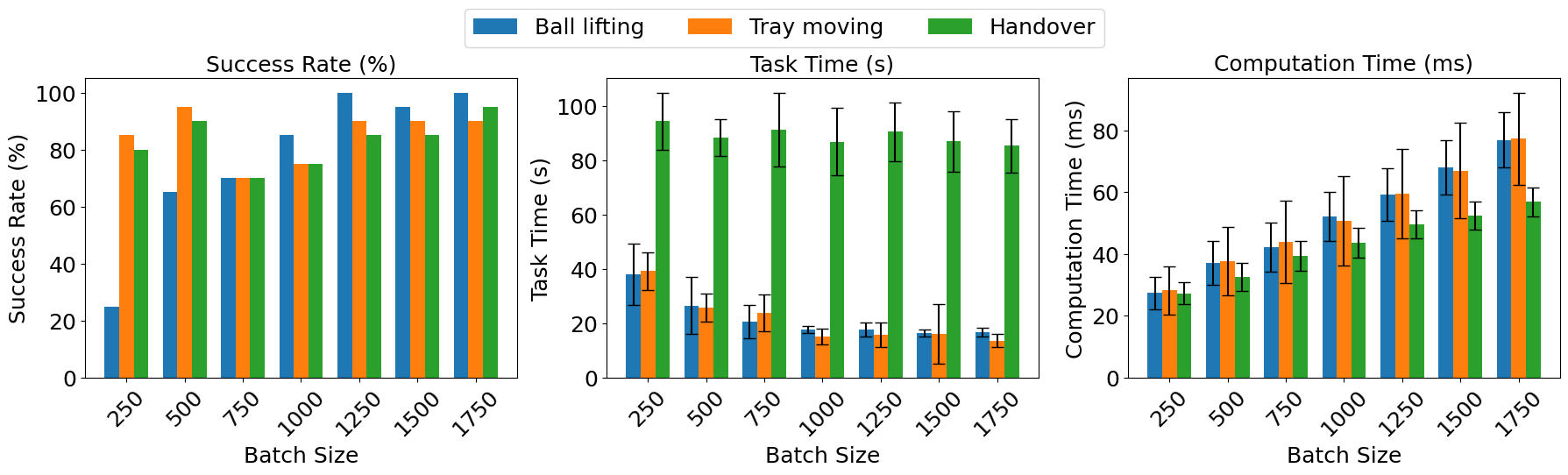}
    \caption{\footnotesize{Statistics of all tasks. (a) Success rate, (b) task time, and (c) computation time with respect to different batch sizes used in Alg.\ref{algo_1}. }}
    \label{fig:stat_all_task}
    \vspace{-0.6cm}
\end{figure*}

\noindent \textbf{Success Metric}
\begin{align}
\text{Success Rate (\%)} = \frac{N_{\mathrm{success}}}{N_{\mathrm{total}}} \times 100
\end{align}
where $N_{\mathrm{success}}$ is the number of successful runs and $N_{\mathrm{total}}$ is the total number of experiments of a task. Failure occurs when there is at least one collision, the task is not completed within 120s, or the ball falls in the ``ball-lifting" task.\\
\noindent \textbf{Mean and Standard Deviation of Computation Time}
\begin{equation}
\begin{aligned}
\bar{t}_{\mathrm{comp}} &= \frac{1}{N_{\mathrm{steps}}} \sum_{m=1}^{N_{\mathrm{steps}}} t_{\mathrm{comp}}^{(m)}, \\
\sigma_{\mathrm{comp}} &= \sqrt{\frac{1}{N_{\mathrm{steps}}} \sum_{m=1}^{N_{\mathrm{steps}}} 
\left( t_{\mathrm{comp}}^{(m)} - \bar{t}_{\mathrm{comp}} \right)^2 }
\end{aligned}
\end{equation}
where $t_{\mathrm{comp}}^{(m)}$ is the computation time for the $m^{th}$ planning step and $N_{\mathrm{steps}}$ is the total number of re-planning needed for the task completion.\\
\noindent \textbf{Mean and Standard Deviation of Task Time}
\begin{equation}
\begin{aligned}
\bar{t}_{\mathrm{task}} &= \frac{1}{N_{\mathrm{runs}}} \sum_{i=1}^{N_{\mathrm{runs}}} t_{\mathrm{task}}^{(i)}, \\
\sigma_{\mathrm{task}} &= \sqrt{\frac{1}{N_{\mathrm{runs}}} \sum_{i=1}^{N_{\mathrm{runs}}} 
\left( t_{\mathrm{task}}^{(i)} - \bar{t}_{\mathrm{task}} \right)^2 }
\end{aligned}
\end{equation}
where $t_{\mathrm{task}}^{(i)}$ is the total time to complete the task in run $i$ and $N_{\mathrm{runs}} = 20$.

\subsubsection*{Experimental Setup}
\noindent A high-fidelity sim-to-real transfer was achieved by meticulously aligning the physical and simulated environments, with the simulation being continuously updated to reflect the real world. The implementation involved the following hardware and data pipeline (Fig.~\ref{fig:real_life_setup}). A unified network was established by connecting both manipulators and a control laptop (PC2) to a single Wi-Fi router using Ethernet, ensuring reliable, low-latency command and feedback. The control laptop was responsible for running the MuJoCo simulation and our proposed planner.
In addition, the control laptop was wirelessly connected to the same Wi-Fi network as the PC1 that controlled the OptiTrack system, which broadcast the reception of the pose data of dynamic objects (obstacles, ball, tray, etc.) through ROS2 topics.
The control laptop subscribed to these topics, integrating the incoming OptiTrack data directly into the simulation by updating the Mocap body features within MuJoCo. This mechanism synchronized the simulation with the physical workspace in real-time.

\vspace{-1em}


\subsection{Qualitative and Quantitative Results}

\noindent Our qualitative and quantitative results are presented in Figures \ref{fig:all_task_snapshots}-\ref{fig:stat_all_task}, respectively. We evaluated our method on three challenging tasks, depicted in Fig.~\ref{fig:all_task_snapshots}(a-d): lifting and transporting a ball over a barrier; (e-h): moving a tray to a goal, by avoiding wall-like obstacles; and (i-l): a hand-over task where one robot passes a cube to another amidst multiple obstacles. The hand-over task was also successfully demonstrated in a real-world experiment, with results available in the supplementary videos.

To benchmark the performance of our approach, we randomized the goal poses in all tasks. The metrics were assessed over multiple batch sizes, i.e., the number of samples used in Alg.\ref{algo_1}, and statistics were collected over 20 runs for each task (Fig.~\ref{fig:stat_all_task}). Note that `Ball lifting' in Fig.~\ref{fig:stat_all_task} refers to the ball lifting and transportation task.

\textbf{Success Rate:} Ball lifting and transporting reached nearly $100\%$ success for batch sizes $\geq1250$. Tray moving maintained $>80\%$ up to 500, dropped at 750, and then recovered with larger sizes. Handover showed a similar trend while peaking at $95\%$ at batch size 1750.

\textbf{Task Time:} Task time decreased with larger batch sizes for ball lifting and tray moving, while handover showed only marginal improvement.

\textbf{Computation Time:} Computation time increased with batch size but stayed below $100\text{ms}$ for all tasks, ensuring real-time applicability; the handover task achieved its peak success with computation just under $60\text{ms}$.

It is worth pointing out that the best success rate presented in Fig.~\ref{fig:stat_all_task} is already comparable to that achieved with best learning approaches in a simpler variant of the tasks in free-space \cite{gkanatsios20253d}. Moreover, the computation time is below several baselines reported in \cite{gkanatsios20253d}.

\subsection{Comparison with \cite{gkanatsios20253d}  }
\noindent We present a preliminary comparison with the work of \cite{gkanatsios20253d}. Since their method was trained in a different simulator, a fair quantitative benchmark is difficult. Instead, we qualitatively demonstrate our improvements on a ball-lifting task from \cite{gkanatsios20253d}, modified with an obstacle requiring a new lifting strategy, recreated in MuJoCo with a Franka-Panda arm.

The qualitative results are summarized in Figure~\ref{fig:teaser}. As shown, the policy from \cite{gkanatsios20253d} is able to partially adapt by navigating around the obstacle to approach the ball. However, it fails when attempting to lift the ball. In contrast, our approach successfully completes the task, with the manipulator lifting the ball while avoiding the obstacle. We repeated this experiment 10 times with a slightly perturbed obstacle position. Our approach succeeded $6$ times as compared to just $2$ for  \cite{gkanatsios20253d}. The comparison videos are shown in the accompanying video. Finally, this experiment also demonstrates that our approach can generalize to different manipulators.  

\section{Conclusions and Future Work}
In this work, we presented a sampling-based optimization framework for complex bimanual manipulation, leveraging a GPU-accelerated physics simulator as a high-fidelity world model. Our approach utilizes a customized Model Predictive Path Integral (MPPI) controller, which embeds a Quadratic Program to ensure smooth, jerk-bounded trajectories. This unique integration facilitates robust exploration in high-dimensional action spaces while preventing undesirable oscillations, a critical challenge in manipulator control. We demonstrated the efficacy and robustness of our method on several challenging, long-horizon tasks—including moving a tray and transporting a ball through cluttered environments—that pose significant generalization difficulties for purely learning-based methods. Our experimental results confirm that the framework achieves high success rates, operates in real-time on commodity GPUs, and enables successful sim-to-real transfer, establishing it as a powerful and generalizable paradigm for solving contact-rich robotic tasks in dynamic settings.

Our work can be extended by distilling the generated optimal trajectories into a policy network, which can then warm-start the optimizer in novel scenarios. Furthermore, our approach could serve as a building block in a TD-MPC framework \cite{hansen2022temporal}, where a learned value function would guide our sampling-based optimizer to further improve performance.

\vspace{-0.5cm}
\bibliography{references}
\bibliographystyle{IEEEtran}

\end{document}